\pdfoutput=1

\documentclass[11pt]{article}

\usepackage[preprint]{acl}

\usepackage{times}
\usepackage{latexsym}

\usepackage[T1]{fontenc}

\usepackage[utf8]{inputenc}

\usepackage{microtype}

\usepackage{inconsolata}

\usepackage{graphicx}

\usepackage{kotex}
\usepackage{color}
\usepackage{makecell}
\usepackage{xurl}
\usepackage{graphicx}
\usepackage{multirow}
\usepackage{subcaption}
\usepackage{booktabs}
\usepackage{xcolor, soul}
\usepackage{amssymb}
\usepackage{bbm}
\usepackage{amsmath}
\usepackage{arydshln}
\usepackage{tcolorbox}
\usepackage{float}

\title{HerO at AVeriTeC: The Herd of Open Large Language Models for Verifying Real-World Claims}

\author{
Yejun Yoon$^{\heartsuit}$~~~~~~Jaeyoon Jung$^{\clubsuit\diamondsuit}$~~~~~~Seunghyun Yoon$^{\spadesuit}$~~~~~~Kunwoo Park$^{\clubsuit\heartsuit}$\\ 
$^{\heartsuit}$Department of Intelligent Semiconductors, Soongsil University\\
$^{\clubsuit}$School of AI Convergence, Soongsil University\\
$^{\diamondsuit}$MAUM AI Inc.\\
$^{\spadesuit}$Adobe Research, USA\\
\texttt{\{yejun0382, jaeyoonskr\}@soongsil.ac.kr}, \texttt{syoon@adobe.com}, \texttt{kunwoo.park@ssu.ac.kr}
}

\newcounter{KWNumberOfComments}
\stepcounter{KWNumberOfComments}

\begin{document}
\maketitle
\begin{abstract}
To tackle the AVeriTeC shared task hosted by the FEVER-24, we introduce a system that only employs publicly available large language models (LLMs) for each step of automated fact-checking, dubbed the \textbf{Her}d of \textbf{O}pen LLMs for verifying real-world claims (\textbf{HerO}). For evidence retrieval, a language model is used to enhance a query by generating hypothetical fact-checking documents. We prompt pretrained and fine-tuned LLMs for question generation and veracity prediction by crafting prompts with retrieved in-context samples. HerO achieved 2nd place on the leaderboard with the AVeriTeC score of 0.57, suggesting the potential of open LLMs for verifying real-world claims. For future research, we make our code publicly available at \url{https://github.com/ssu-humane/HerO}.
\end{abstract}

\section{Introduction}

Automated fact-checking is a task that predicts a claim's veracity by referring to pieces of evidence~\cite{guo-etal-2022-survey}. Claim verification requires the retrieval of relevant information from a reliable document collection and the decision on whether the claim is supported by the known relevant information. Early research attempted to automate the fact-checking process by generating synthetic claims based on Wikipedia documents~\cite{Thorne18Fact,Aly21Feverous} or collecting manually verified claims by human experts~\cite{wang-2017-liar,augenstein-etal-2019-multifc}. However, most datasets suffer from critical issues such as context dependence, evidence insufficiency, and temporal leaks; these limitations made the resulting systems less applicable to the verification of real-world claims. In light of this, a recent study proposed a dataset called AVeriTeC~\cite{averitec}. They addressed the limitations by conducting fine-grained crowdsourced annotations for the fact-checking process.

This paper describes our system for the AVeriTeC shared task hosted by the FEVER-24 workshop~\cite{averitec-shared-task}. Motivated by the recent advancements in large language models, we introduce a fact-checking system that utilizes LLMs for each step of evidence-based fact verification: evidence retrieval, question generation, and veracity prediction. Our system, the \textbf{Her}d of \textbf{O}pen LLMs for verifying real-world claims (\textbf{HerO}), employs publicly available LLMs without using proprietary LLMs, to ensure the transparency of the system. HerO achieved 2nd place in the shared task with an AVeriTeC score of 0.57. Given that the winning system used gpt-4o~\cite{averitec-shared-task}, HerO's competitive performance imply the potential of open LLMs for verifying real-world claims. 

\begin{figure*}[ht]
    \centering
    \includegraphics[width=.99\linewidth]{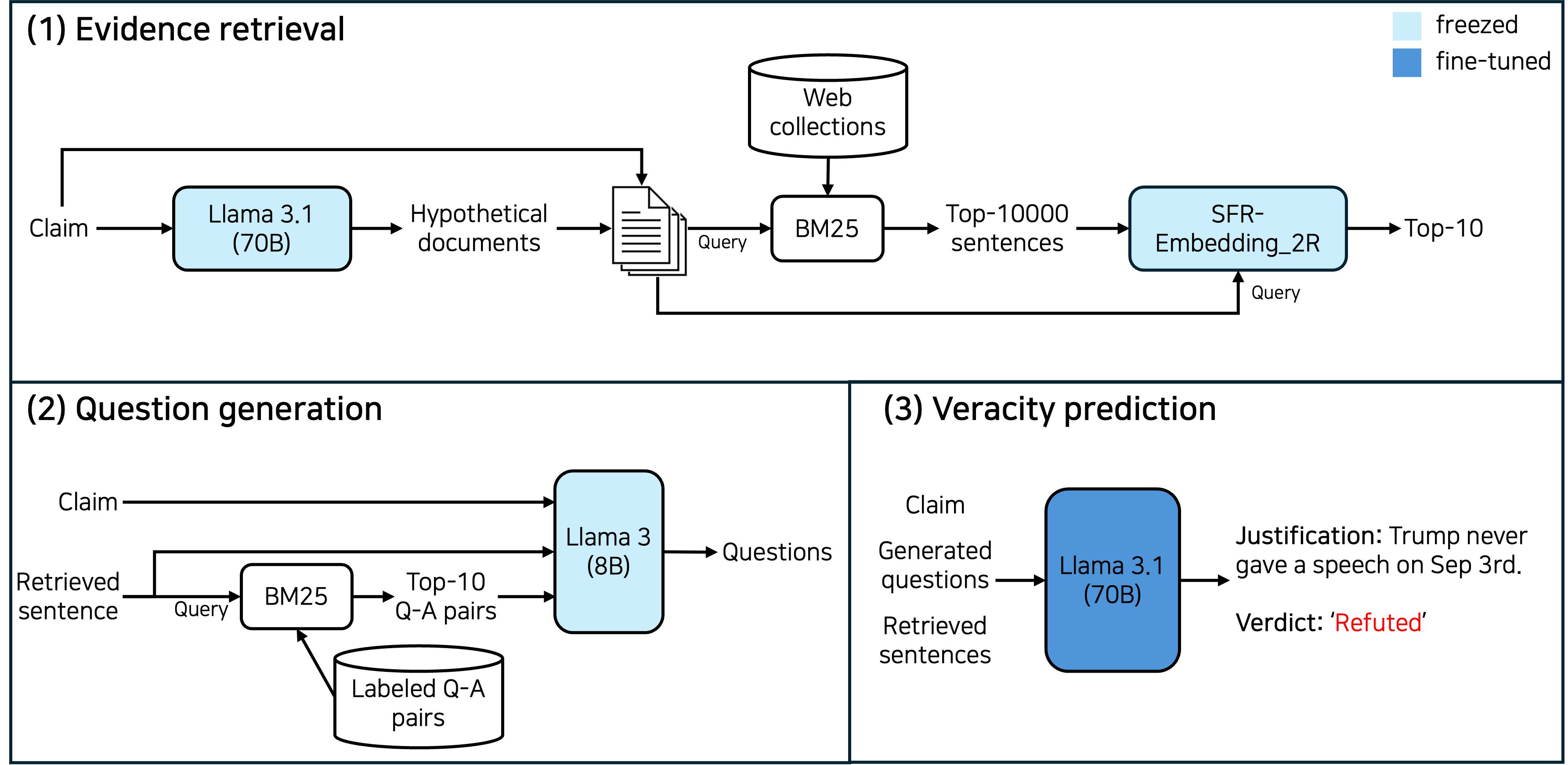}
    \caption{Inference pipeline of our system}
    \label{fig:pipeline}
\end{figure*}

\begin{table*}[ht]
\small
\centering
\resizebox{.99\linewidth}{!}{
\begin{tabular}{ccccccc}
\toprule
\multirow{2}{*}{System} & \multicolumn{2}{c}{Evidence Retrieval} & \multirow{2}{*}{Question Generation} & \multirow{2}{*}{Reranking} & \multirow{2}{*}{Veracity Prediction}\\
& Query & Model &  \\\midrule
 Baseline & Claim & BM25 & Bloom-7b & BERT-base & BERT-base \\
 HerO & \makecell{HyDE-FC\\(Llama-3.1-70b)} & \makecell{BM25\\w/ SFR-embedding-2} & Llama-3-8b & - & Llama-3.1-70b  \\\bottomrule
\end{tabular}
}
\caption{Model configurations}
\label{tab:dev_results_system}
\end{table*}

\section{Related Work}
LLMs have achieved remarkable success in natural language understanding and generation~\cite{gpt3, thoppilan2022lamda, achiam2023gpt}. While major tech companies primarily drove the initial success, they only provided limited access to the model through an API. On the other hand, some research groups have attempted to develop open LLMs to facilitate open research. While the performance of the initial models was unsatisfactory~\cite{zhang2022opt,le2023bloom}, recent models are on par with closed models and even outperform them in certain categories~\cite{jiang2023mistral,dubey2024llama}. 

\section{Task Definition}

The AVeriTeC shared task aims to develop a fact-checking system that verifies real-world claims by retrieving evidence from the web. To verify a given claim, the system first needs to retrieve relevant information from the web documents (evidence retrieval). For each of the collected evidence, the system may generate questions that can help verify the claim (question generation) or choose not to. The last step of the fact-checking is to verify the claim by referring to the collected information (veracity prediction). The final verdict is a four-class variable: supported, refuted, not enough evidence, or conflicting evidence/cherry-picking. Each system is evaluated using three metrics, where a higher value indicates a better score. Two metrics are the Hungarian METEOR score\footnote{The score uses the Hungarian algorithm~\cite{hungarian} to find optimal matching pairs and evaluates them with the METEOR score~\cite{meteor}.} to assess the quality of questions (Q score) and question-answer pairs (Q$+$A score), respectively. The overall accuracy is measured by the AVeriTeC score. Details about the task, dataset, and evaluation metrics can be found in \citet{averitec} and \citet{averitec-shared-task}.

\section{Our System}

This section describes our fact-checking system, the \textbf{Her}d of \textbf{O}pen LLMs for verifying real-world claims (\textbf{HerO}). Inspired by the recent progress of open LLMs~\cite{jiang2023mistral, dubey2024llama}, we only employ open LLMs for our system without using proprietary LLMs, such as gpt~\cite{gpt3} and gemini~\cite{team2023gemini}. Table~\ref{tab:dev_results_system} presents HerO's model configurations in comparison to the baseline system~\cite{averitec}. The inference pipeline of our system is illustrated in Figure~\ref{fig:pipeline}. We use web documents provided along with the dataset as the knowledge store.

\subsection{Evidence Retrieval}

The first step aims to retrieve relevant sentences from the knowledge store to verify a given claim. Inspired by previous research on generative retrieval methods~\cite{gao-etal-2023-precise,wang-etal-2023-query2doc}, we utilize an instruction-following LM to generate hypothetical fact-checking documents to augment a retrieval query. For the rest of this paper, we call this approach HyDE-FC, which stands for Hypothetical Document Embedding for Fact-Checking. 

Given a claim $c$, we generate a set of hypothetical fact-checking documents $D=\{d_{1},\dots,d_{N}\}$ by prompting an instruction-following language model $f(\cdot)$ using $c$ as an in-context sample. The used prompt for HyDE-FC is shown in Figure~\ref{fig:hyde-fc-prompt}. We repeat the sampling process until obtaining $N$ different documents.

\begin{figure}[t]
\begin{tcolorbox}[colback=white, fontupper=\small]
\textbf{Please write a fact-checking article passage to support, refute, indicate not enough evidence, or present conflicting evidence regarding the claim.}\\
\textbf{Claim}: \textit{Hunter Biden had no experience in Ukraine or in the energy sector when he joined the board of Burisma.}\\
\textbf{Passage}: \textcolor{blue}{While Hunter Biden did not have direct experience in the energy sector or Ukraine before joining the board of Burisma, he did have ...}
\end{tcolorbox}
\centering
\caption{An example of the instruction prompt used for HyDE-FC and its output. The bold text is the instruction, the italic text is a claim, and the blue text indicates the model output.}
\label{fig:hyde-fc-prompt}
\end{figure}

\begin{figure}[t!]
\begin{tcolorbox}[colback=white, fontupper=\small]
\textbf{Your task is to generate a question based on the
given claim and evidence. The question should
clarify the relationship between the evidence and
the claim}\\\\
\textbf{Example 1}:\\
\textbf{Claim}: \textcolor{gray}{U.S. aid dollars sent to Ukraine under Biden’s supervision went toward Burisma, where Biden’s son Hunter was a board member.}\\
\textbf{Evidence}: \textcolor{gray}{Hunter Biden was appointed to the board of Burisma.}\\
\textbf{Question}: \textcolor{gray}{Was Hunter Biden a board member of Ukrainian energy company 'Burisma'?}\\
\textbf{...}\\
\textbf{Example 10}:\\
\textbf{Claim}: \textcolor{gray}{Hunter Biden was paid $3 million plus $183,000 a month to be a board member of a company that a lot of people said was corrupt.}\\
\textbf{Evidence}: \textcolor{gray}{Burisma Holdings, Ukraine’s largest private gas producer, has expanded its Board of Directors by bringing on Mr. R Hunter Biden as a new director.}\\
\textbf{Question}: \textcolor{gray}{What company is Hunter Biden a member of the board?}\\\\
\textbf{Now, generate a question that links the following
claim and evidence:}\\\\
\textbf{Claim}: \textit{Hunter Biden had no experience in Ukraine or in the energy sector when he joined the board of Burisma.}\\
\textbf{Evidence}: \textcolor{gray}{In 2014, Hunter Biden was appointed to the board of Burisma Holdings, a Ukrainian energy company. He was reportedly paid \$50,000 a month to work in an industry in which he had no previous experience.}\\
\textbf{Question}: \color{blue}What was Hunter Biden's background or experience in the energy sector before joining the board of Burisma Holdings in 2014?\\
\end{tcolorbox}
\centering
\caption{An example of instruction prompt and its output for question generation. The bold text indicates the instruction, the italic text is a claim, the gray text is retrieved in-context samples, and the blue text indicates the model output.}
\label{fig:question-prompt}
\end{figure}

Using the claim and generated documents, our retrieval pipeline employs a two-step hybrid approach that incorporates spare and dense retrieval methods. The first step is to retrieve relevant documents by BM25~\cite{bm25}. We concatenate the claim $c$ and each document in $D$ for building the query document $q$. The sparse vector for $q$ is used to retrieve the top 10,000 relevant sentences from the knowledge store. The second step is to re-rank the 10,000 sentences by the dense retrieval method to decide the top-10 evidence candidates. The query vector $v_q$ is obtained by averaging the embedding vectors for the claim $c$ and every document in $D$ by the equation \ref{eq2}, \begin{equation}\label{eq2}
v_{q} = \frac{1}{N+1}[\sum\limits^{N}_{k=1}g(d_{k})+g(c)]
\end{equation}
where $g$ is an embedding method. 

Our best model uses llama-3.1-70b~\cite{dubey2024llama} for $f$ and SFR-embedding-2~\cite{SFR-embedding-2} for $g$. $N$ is set as 8.

\subsection{Question Generation}

The next step is to generate verifying questions, each of which the corresponding answer could be a retrieved sentence. We employ an instruction-following LM to generate questions for each piece of evidence. The used prompt is shown in Figure~\ref{fig:question-prompt}. We improve the baseline prompt~\cite{averitec}, which takes each evidence and relevant question-answer pairs from the labeled set by BM25 as in-context examples, by including a corresponding claim.

\subsection{Veracity Prediction}

\begin{figure}[ht]
\begin{tcolorbox}[colback=white, fontupper=\small]
\textbf{Your task is to predict the verdict of a claim
based on the provided question-answer pair evidence. 
The possible labels are: ’Supported’, ’Refuted’, ’Not Enough Evidence’, ’Conflicting Evidence/Cherrypicking’. Justify your answer using
the provided evidence and select the correct label.}\\\\
\textbf{Claim}: \textit{Hunter Biden had no experience in Ukraine or in the energy sector when he joined the board of Burisma.}\\\\
\textbf{Q1}: \textcolor{gray}{What was Hunter Biden's background or experience in the energy sector before joining the board of Burisma Holdings in 2014?}\\
\textbf{A1}: \textcolor{gray}{In 2014, Hunter Biden was appointed to the board of Burisma Holdings, a Ukrainian energy company. He was reportedly paid \$50,000 a month to work in an industry in which he had no previous experience.}\\
\textbf{...}\\
\textbf{Q10}: \textcolor{gray}{Did Hunter Biden have any relevant experience in Ukraine or the energy sector before joining the board of Burisma?}\\
\textbf{A10}: \textcolor{gray}{What this is all about: From the start of the inquiry, Republicans have pointed out that Hunter Biden did not have any experience in corporate governance or in the energy sector before taking the job at Burisma.}\\

\textbf{Justification}: \textcolor{blue}{No former experience stated.}\\
\textbf{Verdict}: \textcolor{blue}{Supported}\\
\end{tcolorbox}
\centering
\caption{An example of instruction prompt and its output for veracity prediction. The bold text indicates the instruction, the italic text is a claim, the gray text is retrieved QA pairs, and the blue text is the model output.}
\label{fig:verdict-prompt}
\end{figure}

We employ an instruction-following LM for veracity prediction. Inspired by a previous study~\cite{CoT}, we devise a prompt that incorporates an annotator's rationale into the veracity prediction. Our best model uses the fine-tuned llama-3.1-70b-it that predicts the veracity label after generating the explanation. The top 10 question-and-answer pairs from the earlier steps are given as in-context samples along with the claim to verify.

\section{Evaluation Experiments}

In this section, we present experimental results to decide the system configuration. 

\subsection{Experimental Setups}

In the comparison experiments, we used the development set to evaluate model performance. In addition to the Q score and Q$+$A score, we employed the Hungarian METEOR score to evaluate the answer quality, denoted as A score. For the comparison experiments, we used the training set for training our models and the development set for the evaluation.  The training and development set were used to train our system for the submission. We used the Adam optimizer with a learning rate 2e-5, batch size 128, and 2 epochs. For LoRA, we set the rank to 128 and alpha to 256.

All the language models used in the experiments are the instruction-tuned version (e.g., llama-3.1-70b-it). For brevity, we omitted `it' in the model identifier for the rest of the paper. For HyDE-FC, we set the LM hyperparameters as follows: maximum number of tokens as 512, temperature as 0.7, and top\_p as 1.0. We used the labeled QA pairs from the training set as a data store to retrieve in-context samples for question generation. We used greedy decoding with a maximum length of 512. When an LM does not produce the verdict label, we repeated the generation with the top-2 sampling.

We ran experiments using three machines. The first has two H100 GPUs (80GB per GPU) and 480GB RAM. The second has eight H100 GPUs with 2TB RAM; the third has four NVIDIA A6000 GPUs (48GB per GPU) and 256GB RAM. The experiments were conducted in a computing environment with the following configuration: Python 3.11.9, PyTorch 2.3.1, Transformers 4.43.4, Axolotl 0.4.1, vLLM 0.5.3, and SentenceTransformers 3.0.1. HerO took approximately 6.6 hours to make 500 predictions for the development set with two H100 GPUs. It took six hours for the evidence retrieval, 25 minutes for the question generation, and 12 minutes to complete the veracity prediction.

\begin{table}[t]
\small
\centering
\begin{tabular}{cccc}
\toprule
Query & Retrieval model & A score \\\midrule
\multirow{3}{*}{Claim} & BM25 & 0.187 \\
& \makecell{BM25\\w/ SFR-embedding-2} & 0.26 \\\midrule

\makecell{HyDE-FC\\(Llama-3-8b)} & \multirow{9}{*}{\makecell{BM25\\w/ SFR-embedding-2}} & 0.2745 \\
\makecell{HyDE-FC\\(Llama-3-70b)} &  & 0.2757 \\
\makecell{HyDE-FC\\(Llama-3.1-8b)} &  & 0.2751 \\
\makecell{HyDE-FC\\(Llama-3.1-70b)} &  & \textbf{0.2801} \\
\makecell{HyDE-FC\\(GPT-4o-mini)} & & 0.2773\\\midrule 
\end{tabular}
\caption{Performance of evidence retrieval methods}
\label{tab:dev_hyde}
\end{table}

\begin{table}[t]
\small
\centering
\begin{tabular}{ccc}
\toprule
Context & Model & Q score \\
\midrule
\multirow{6}{*}{Retrieved sentences} & Baseline & 0.2404 \\
& Llama-3-8b & 0.4210 \\
& Llama-3-70b & 0.4175 \\
& Llama-3.1-8b & 0.4212 \\
& Llama-3.1-70b & 0.4259 \\
& GPT-4o-mini & 0.4054 \\
\midrule
\multirow{4}{*}{\makecell{Retrieved sentences\\w/ Claim}} & Llama-3-8b & \textbf{0.4938} \\
 & Llama-3-70b & 0.4789 \\
 & Llama-3.1-8b & 0.4855 \\
 & Llama-3.1-70b & 0.4881 \\
\bottomrule
\end{tabular}
\caption{Performance of question generation methods}
\label{tab:dev_qg}
\end{table}

\subsection{Experimental Results}

\paragraph{Evidence Retrieval} 
We present evidence retrieval results on the AVeriTeC development set in Table~\ref{tab:dev_hyde}. We relied on the A score as the primary metric to identify a model that can retrieve sentences that are similar to the annotated evidence. 

We made three observations. First, when a claim was used as a query verbatim, applying SFR-embedding-2 to the re-ranking step boosted the performance by the A score of 0.073. Second, augmenting a query by the hypothetical document generation increased the performance. The best model, HyDE-FC with llama-3.1-70b, achieved an A score of 0.2801, 0.02 greater than the claim-only approach. Third, gpt-4o-mini was close to but slightly worse than the best open model when being used for HyDE-FC. Accordingly, HerO uses the two-step approach where SFR-embedding-2 re-ranks the top 10,000 sentences obtained by BM25; llama-3.1-70b is used to generate hypothetical fact-checking documents to augment the query.

\paragraph{Question Generation} 
We present evaluation results of question generation methods in Table~\ref{tab:dev_qg}. We fixed the evidence retrieval method as the best approach to assess the effects of question generation methods. The Q score was used as a primary evaluation metric for question generation. 

We made three observations. First, all the llama models achieved better Q scores than the baseline and gpt-4o-mini. Second, using the claim as an additional in-context sample boosted the generation performance significantly. The llama-3-8b model with the claim achieved a Q score of 0.4938, 0.0728 greater than its counterpart. Third, among the llama models that only use retrieved sentences as in-context samples, the latest and largest model (llama-3.1-70b) achieved the best score. However, llama-3-8b achieved the best score with the claim. Accordingly, HerO uses llama-3-8b to generate questions.

\paragraph{Veracity Prediction}
We compared veracity prediction methods using the best evidence retrieval and question generation pipelines. We evaluated three LLM-based methods: in-context learning with ten examples, instruction fine-tuning by LoRA~\cite{lora}, and fine-tuning the whole parameters. Table~\ref{tab:dev_vp} shows the results. When in-context learning was used without parameter updates, the llama models outperformed gpt-4o-mini. The most significant performance gap was an accuracy of 0.14 and an AVeriTeC score of 0.112. Furthermore, the performance was boosted by instruction fine-tuning approaches. The llama-3.1-70b with the full fine-tuning approach achieved the highest AVeriTeC score of 0.578, which is the veracity prediction module for HerO.

\subsection{Test Set Results} 

Table~\ref{tab:test_results} shows how HerO performs in the test set in comparison to the baseline and other competitive models. TUDA\_MAI\_0 achieved the best AVeriTeC score of 0.63, followed by HerO (0.57) and CTU AIC (0.5). Their performance gap with the existing baseline was significant. HerO achieved the best Q and Q+A scores among the top 3 models, suggesting that our question-generation approach is strong. Since HerO's performance gap with the winning system was smaller for the Q+A score than for the Q score, we suspected that our retrieval system is on par with but slightly worse than theirs. The answer score employed in our experiment could help better understand what is attributed to the performance, either retrieval or question generation.

\begin{table}[t]
\small
\centering
\resizebox{.99\linewidth}{!}{
\begin{tabular}{cccc}
\toprule
Method & Model & Accuracy & AVeriTeC score \\
\midrule
\multirow{3}{*}{\makecell{In-context\\learning}} & Llama-3-70b & 0.628 & 0.494 \\
& Llama-3.1-70b & 0.54 & 0.422 \\
& Gpt-4o-mini & 0.488 & 0.382 \\
\midrule
\multirow{2}{*}{LoRA} & Llama-3-70b & 0.724 & 0.556 \\
& Llama-3.1-70b & 0.704 & 0.55 \\
\midrule
\multirow{2}{*}{Fine-tuning} & Llama-3-70b & 0.746 & 0.57 \\
& Llama-3.1-70b & \textbf{0.752} & \textbf{0.578} \\
\bottomrule
\end{tabular}
}
\caption{Performance of veracity prediction methods}
\label{tab:dev_vp}
\end{table}

\begin{table}[t]
\small
\centering
\resizebox{.99\linewidth}{!}{
\begin{tabular}{cccc}
\toprule
System &  Q score & Q+A score & AVeriTeC score\\\midrule
TUDA\_MAI\_0 & 0.45 & 0.34 & \textbf{0.63} \\
HerO & \textbf{0.48} & \textbf{0.35} & 0.57 \\
CTU AIC & 0.46 & 0.32 & 0.5 \\
Baseline & 0.24 & 0.2 & 0.11 \\
\bottomrule
\end{tabular}
}
\caption{Test set results}
\label{tab:test_results}
\end{table}

\section{Conclusion}
To tackle the AVeriTeC shared task hosted by the FEVER-24, we developed HerO, a fact-checking system that employs publicly available large language models for each step of automated fact-checking: evidence retrieval, question generation, and veracity prediction. Our system achieved 2nd place in the shared task, supporting the effectiveness of open LLMs for verifying real-world claims. We release our code publicly for future research. 

\section*{Acknowledgments}
This research was supported by the MSIT(Ministry of Science and ICT), Korea, under the Graduate School of Metaverse Convergence support (IITP-2024-RS-2024-00430997) and Innovative Human Resource Development for Local Intellectualization (IITP-2024-RS-2022-00156360) programs, supervised by the IITP(Institute for Information \& Communications Technology Planning \& Evaluation). We used an equipment supported by the NIPA (National IT Promotion Agency) under the high performance computing support program. The title and system name are hommage to research on open language models (to list a few, \citet{jiang2023mistral}, \citet{SFR-embedding-2}, and \citet{dubey2024llama}), which made possible the development of our fact-checking system. Yejun Yoon and Jaeyoon Jung contributed to this work equally as co-first authors. Kunwoo Park is the corresponding author.

\bibliography{acl_latex}

\end{document}